\documentclass[11pt,a4paper]{article}

\usepackage[utf8]{inputenc}
\usepackage[T1]{fontenc}
\usepackage[margin=2.5cm]{geometry}
\usepackage{graphicx}
\usepackage{amsmath,amssymb}
\usepackage{booktabs}
\usepackage{threeparttable}
\usepackage{enumitem}
\usepackage{float}
\usepackage{caption}
\usepackage[numbers,sort&compress]{natbib}
\usepackage[hidelinks]{hyperref}

\newcommand{\amil}{AMIL}
\newcommand{\vote}{majority voting}
\newcommand{\pp}{\,pp}
\newcommand{\bnet}{BarkNet 1.0}
\newcommand{\ntrees}{998}
\newcommand{\nclasses}{20}
\newcommand{\htag}[1]{\textnormal{\small(\textbf{#1})}}

\title{\bfseries Toward an Optimal Configuration for Tree Species
Classification from Bark Images:\\[2pt] A Leakage-Safe Ablation Study}

\author{Maeve C\^ot\'e\thanks{Independent researcher, Montr\'eal, Canada;
\texttt{maeve04.cote@gmail.com}}}

\date{\today}

\begin{document}
\maketitle

\begin{abstract}
\noindent
\textit{Background:} The automatic identification of tree species from bark
photographs is of practical interest for forestry inventory and for the
wood-processing industry, yet the numerous design choices that govern the
accuracy of deep-learning approaches have not been studied systematically, and
some prior results may be inflated by the leakage of images of the same trunk
across training and test partitions.
\textit{Methods:} A systematic ablation study is conducted on the public \bnet{}
dataset (\nclasses{} species) under the established tree-level, leakage-safe
partitioning scheme. Using ConvNeXt-V2 backbones, the following factors are varied
in turn and evaluated by five-fold cross-validation: the patch size at which bark is
sampled (five sizes), the strategy by which per-patch predictions are aggregated
into an image-level decision, the training regime (single- versus two-stage), the
backbone capacity, the whole-image baseline, and the number of patches available at
inference. A matched image-level split of the same data measures the effect of
same-trunk leakage.
\textit{Results:} Per-patch accuracy rises monotonically with patch size (from
$75.2\%$ at $160$\,px to $89.3\%$ at $512$\,px), and image-level accuracy rises in
step, plateauing near $94.9\%$ around $512$\,px. Learned aggregation exceeds majority
voting by at most $0.7$\,pp, an advantage that shrinks as patches grow and the
attention distribution becomes near-uniform. Soft voting and uniform (mean) pooling of
the same embeddings do no better than hard voting, so the gain comes from the learned
attention weights alone and vanishes at $512$\,px. Backbone capacity (pico to tiny) alters
accuracy by less than a point, and two-stage training does not offer an advantage
over one-stage training. Whole-image classification ($91.9\%$) trails patch-based
aggregation despite its larger field of view, since it admits no bag to average over;
and accuracy increases with the number of patches available at inference, saturating
around $64$ patches. Under an image-level split of the same data, the recommended
configuration reaches $99.9\%$ ($+5.0$\,pp) and the whole-image baseline $99.2\%$
($+7.3$\,pp): an inflation comparable to the entire effect of patch size and several
times larger than any other design effect studied.
\textit{Conclusions:} The study yields a simple recommended configuration: patches
as large as possible while still leaving enough patches to average over (around
$512$\,px here), with a nano backbone. It also quantifies the second-order role of
learned aggregation and model capacity, providing a reproducible reference point for
practitioners and future research.
\end{abstract}


\section{Introduction}
\label{sec:intro}

The identification of tree species is a routine but labour-intensive task in forest
inventory, ecological monitoring, and the grading and pricing of timber in the
wood-processing industry, and automating it with computer vision has attracted
sustained interest over the past decade~\cite{waldchen2018trends,wei2022fgia}. Most
automated systems identify plants from leaves, flowers, or fruit~\cite{waldchen2018trends},
which are seasonal and frequently out of reach on a standing trunk. Bark, by
contrast, is visible throughout the year and remains accessible on both standing and
felled trunks, and its texture varies comparatively little with season; these
properties have made it an attractive cue for automated
identification~\cite{fiel2011bark,cui2023convnext,boudra2018srbp,carpentier2018barknet}.

Early approaches to bark recognition relied on hand-crafted texture descriptors,
such as local binary pattern variants and statistical radial binary
patterns~\cite{fiel2011bark,boudra2018srbp,fekriershad2020bark} and multispectral
Markovian texture models~\cite{remes2019bark}. The feasibility of learning such
features directly was established by Carpentier et al.~\cite{carpentier2018barknet},
who released the \bnet{} dataset and reported strong accuracy by classifying image
crops with a convolutional network and aggregating the per-crop predictions through
hard \vote{}; Wu et al.~\cite{wu2021deepbarkid} later kept the same tree-level split
and distilled the network into a compact model. Subsequent work has pursued higher accuracy with more capable
backbones; Cui et al.~\cite{cui2023convnext}, for example, reported near-saturated
single-image accuracy on an extended dataset using a ConvNeXt
architecture~\cite{liu2022convnet}. Beyond \bnet{}, larger bark benchmarks have since
been released, including Bark-101~\cite{ratajczak2019bark}, BarkVN-50~\cite{truonghoang2020barkvn50,faizal2022barkvn},
CentralBark~\cite{warner2024centralbark}, and the recent BarkBD~\cite{ali2026barknetlite}.

Viewed more abstractly, bark recognition is an instance of fine-grained visual
categorization~\cite{wei2022fgia}: the classes are visually similar sub-categories
separated by subtle, texture-like cues rather than by object shape. It also shares its
computational structure with a broader family of tasks in which a single image is too
large or too heterogeneous to classify directly and is instead tiled into patches
whose predictions are aggregated into an image-level decision. This
tile-and-aggregate paradigm is central to computational pathology, where gigapixel
whole-slide images are split into thousands of patches and combined by multiple
instance learning~\cite{ilse2018attention,campanella2019clinical,lu2021clam,li2021dsmil,zhang2022dtfd},
and it recurs in wood identification, where board or specimen images are classified patch
by patch and the predictions combined by voting~\cite{hafemann2014forest,wu2021woodlongitudinal}. The design questions examined here,
namely how large each tile should be, how many are needed, and how their predictions
should be combined, are therefore not specific to bark, and the conclusions may
inform other tile-based fine-grained recognition tasks.

Despite this progress, two issues limit the practical value of the existing
literature. First, the design choices that determine accuracy (the resolution at
which bark is sampled, the manner in which patch-level evidence is combined, the
training procedure, the capacity of the backbone, and the number of patches
required at inference) have largely been adopted by convention. Individual choices
have been studied in isolation, for instance patch extraction and majority-voting
rules for bark by Misra et al.~\cite{misra2020patch}, segment size and overlap for bark by
Surendran et al.~\cite{surendran2025forestry}, and patch size for wood by Liu et
al.~\cite{liu2024cites}, but they have not been compared
jointly and on a common footing, so no consolidated guidance exists as to the
configuration a practitioner should adopt. Second, when a dataset contains several
photographs of the same trunk, partitioning the data at the level of individual images
rather than at the level of trees allows near-identical views of a trunk to appear in
both training and test sets; the resulting leakage can inflate the reported accuracy
and obscure the true difficulty of the task, all the more so because bark texture is
distinctive enough to re-identify an individual trunk~\cite{robert2020barkreid}. This is a
failure mode documented across many machine-learning-based
sciences~\cite{kaufman2011leakage,kapoor2023leakage}, including tile-based
pathology, where tiles of one patient on both sides of the split inflate
scores~\cite{bussola2021tiles}, and tree-species mapping from drone
imagery~\cite{kattenborn2022spatial}; a tree-disjoint benchmark has recently been proposed
for wood images for the same reason~\cite{casanova2026woodbrcross}.
On \bnet{}, Carpentier et al.~\cite{carpentier2018barknet} adopted a tree-level split,
which we retain; several more recent studies instead split by
image~\cite{cui2023convnext,ali2026barknetlite}, and the authors of the latter note that
their figures are consequently an upper bound.

Accordingly, the present study does not seek to introduce a new architecture or to
maximise a single headline figure. Instead, a systematic and leakage-safe ablation
study is conducted, in which each principal design choice is varied in isolation
and its effect on accuracy is quantified under a common, tree-level evaluation
protocol. The aim is to establish which configuration is most effective for
bark-based species classification, and by how much each factor matters, so that the
result may serve as a dependable starting point for both industrial deployment and
future research.

The contributions of this work are as follows:
\begin{enumerate}[leftmargin=*]
\item A systematic five-fold ablation conducted under the tree-level, leakage-safe
evaluation protocol of Carpentier et al.~\cite{carpentier2018barknet}, with all
results reported as means and standard deviations of accuracy and macro-$F_1$,
together with a direct measurement of the accuracy inflation that image-level splits
produce on the same data.
\item A systematic ablation over the principal design choices (patch size,
aggregation strategy, training regime, backbone capacity, the whole-image baseline,
and test-time bag size), quantifying the contribution of each to overall accuracy.
\item A practical recommended configuration for deployment, derived from the
ablation and supported by an attention-concentration diagnostic and an analysis of
the learned attention.
\end{enumerate}

\section{Materials and Methods}
\label{sec:methods}

\subsection{Dataset}
\label{sec:data}
The public \bnet{} dataset~\cite{carpentier2018barknet} is used throughout. It
comprises photographs of trees belonging to 23 species, acquired with consumer
smartphones under field conditions. Because several photographs were captured for
each trunk, the dataset is well suited to a leakage-safe, tree-level evaluation but
requires care in partitioning (Section~\ref{sec:split}). The \nclasses{} species
with a sufficient number of trees are retained, and the three rarest species (ERB,
PEG, PID) are excluded, yielding \ntrees{} trees and 23{,}271 images (a mean of 23.3
images per tree); this is the full retained set, with no further filtering, and the
per-image cross-validation fold assignments are released with the code to allow exact
reproduction. The resulting class distribution is summarised in
Table~\ref{tab:dataset}.

\begin{table}[H]
\centering
\caption{Class distribution of the \nclasses{} retained species in \bnet{}
(trees, images, and 224\,px patches).}
\label{tab:dataset}
\small
\begin{tabular}{llrrrr}
\toprule
\textbf{Code} & \textbf{Species (Latin)} & \textbf{Trees} & \textbf{Images}
 & \textbf{Patches} & \textbf{Patches/img}\\
\midrule
MEL & \textit{Larix laricina}       &  77 & 1{,}874 & 574{,}385 & 306.5\\
CHR & \textit{Quercus rubra}        & 109 & 2{,}724 & 395{,}026 & 145.0\\
ERS & \textit{Acer saccharum}       &  81 & 1{,}911 & 332{,}432 & 174.0\\
PET & \textit{Populus tremuloides}  &  58 & 1{,}037 & 314{,}582 & 303.4\\
FRA & \textit{Fraxinus americana}   &  61 & 1{,}472 & 282{,}180 & 191.7\\
ERR & \textit{Acer rubrum}          &  64 & 1{,}676 & 264{,}115 & 157.6\\
EPN & \textit{Picea mariana}        &  44 &   885 & 220{,}619 & 249.3\\
BOJ & \textit{Betula alleghaniensis}&  43 & 1{,}255 & 199{,}796 & 159.2\\
EPO & \textit{Picea abies}          &  72 & 1{,}324 & 186{,}426 & 140.8\\
BOP & \textit{Betula papyrifera}    &  32 & 1{,}285 & 184{,}384 & 143.5\\
SAB & \textit{Abies balsamea}       &  41 &   922 & 151{,}021 & 163.8\\
OSV & \textit{Ostrya virginiana}    &  29 &   612 & 146{,}457 & 239.3\\
ORA & \textit{Ulmus americana}      &  25 &   767 & 145{,}963 & 190.3\\
PRU & \textit{Tsuga canadensis}     &  45 &   986 & 143{,}715 & 145.8\\
PIB & \textit{Pinus strobus}        &  39 & 1{,}023 & 138{,}035 & 134.9\\
HEG & \textit{Fagus grandifolia}    &  41 &   840 & 128{,}342 & 152.8\\
EPR & \textit{Picea rubens}         &  27 &   740 & 122{,}009 & 164.9\\
THO & \textit{Thuja occidentalis}   &  38 &   746 & 105{,}487 & 141.4\\
EPB & \textit{Picea glauca}         &  44 &   596 & 102{,}438 & 171.9\\
PIR & \textit{Pinus resinosa}       &  29 &   596 &  78{,}793 & 132.2\\
\midrule
\textbf{Total} & \textbf{\nclasses{} species} & \textbf{998} & \textbf{23{,}271}
 & \textbf{4{,}216{,}205} & \textbf{181.2}\\
\bottomrule
\end{tabular}
\end{table}

\subsection{Leakage-Safe, Tree-Level Partitioning}
\label{sec:split}
To prevent the leakage of near-identical views across partitions, the data are
divided at the level of trees rather than images: every photograph of a given tree
is assigned in its entirety to a single partition. This is the same tree-level split
adopted by Carpentier et al.~\cite{carpentier2018barknet} on \bnet{}, retained here so
that our results remain directly comparable to theirs and free of same-trunk leakage.
This entity-level (group-aware) partitioning is the standard safeguard against the leakage that arises when
correlated images are divided independently, a failure mode documented across many
machine-learning-based sciences~\cite{kapoor2023leakage}, recommended for hierarchically
structured ecological data~\cite{roberts2017cv}, and mirrored by
patient-level splitting in computational pathology~\cite{campanella2019clinical,bussola2021tiles}.
Five-fold cross-validation is performed at the tree level, and all reported
quantities are means and standard deviations across the five folds. Because the folds
partition the trees exhaustively and disjointly, every image is used for evaluation
exactly once, so pooled test predictions sum to the dataset size. The protocol yields
accuracy estimates that are lower, but more faithful, than those obtained under
image-level partitioning, and it is applied uniformly to every configuration examined
in this study so that the ablations remain directly comparable.

\paragraph{Image-level comparison split.} To measure the effect of leakage directly, a
second partition is used for one experiment only (Section~\ref{sec:res-leakage}): images
are assigned to five folds by stratified random sampling within each species, ignoring
tree identity, as in an image-level split. Under this partition, $99.9$--$100\%$ of test images
in each fold belong to a tree that also contributes images to the training folds. The whole-image
baseline and the recommended configuration (nano, $512$\,px, \amil{}) are retrained
under this partition with every other setting unchanged, including hyperparameters and
seeds, so that any difference is attributable to the partition alone.

\subsection{Patch Extraction}
\label{sec:patches}
Each photograph is divided into square patches of side $s$ pixels using a
minimal-overlap tiling that covers the full trunk region. Five sampling resolutions
are considered, $s \in \{160, 224, 288, 384, 512\}$. A photograph is treated as a
\emph{bag} of patches, and each patch is an \emph{instance} within that bag; this
bag-of-patches formulation mirrors the tiling of gigapixel images into instances in
computational pathology~\cite{campanella2019clinical,lu2021clam}. For reference,
Carpentier et al.~\cite{carpentier2018barknet} extracted $224$\,px crops from images
downsampled to half of their native resolution (equivalent to roughly $448$\,px
patches at full resolution), both to speed up loading and processing and to account
for the Bayer colour-filter array, which samples each colour channel at only every
other pixel. It is noted at the outset that patch size and bag size are confounded: at
a fixed image resolution, larger patches yield fewer but individually more
informative instances per bag. This confound is addressed explicitly in
Section~\ref{sec:decouple}.

\subsection{Backbone and Aggregation Model}
\label{sec:model}
A ConvNeXt-V2 backbone~\cite{liu2022convnet,woo2023convnextv2}, initialised from
ImageNet-pretrained weights~\cite{deng2009imagenet}, is used as the per-patch feature
extractor. Three capacities are compared, ConvNeXt-V2 pico, nano, and tiny, to
characterise the effect of model size (Section~\ref{sec:res-backbone}); unless stated
otherwise, results are reported for the nano backbone. For the learned-aggregation
configuration, all $N$ patches of a bag are passed through the backbone $f_\theta$ to
obtain embeddings $\mathbf{h}_i = f_\theta(x_i)$, which are combined by
attention-based multiple instance learning~\cite{ilse2018attention}, an approach
widely used to aggregate patch-level predictions over gigapixel whole-slide images in
computational pathology~\cite{campanella2019clinical,lu2021clam}. An attention module
assigns each patch a weight, and the bag embedding is formed as the weighted sum of
the patch embeddings,
\begin{equation}
\alpha_i = \frac{\exp\!\big(\mathbf{w}^\top \tanh(V \mathbf{h}_i)\big)}
{\sum_{j=1}^{N}\exp\!\big(\mathbf{w}^\top \tanh(V \mathbf{h}_j)\big)},
\qquad
\mathbf{z} = \sum_{i=1}^{N} \alpha_i \, \mathbf{h}_i .
\end{equation}
A shared linear head $g(\cdot)$ maps the bag embedding to image-level logits,
$\hat{y} = g(\mathbf{z})$. The attention aggregation is one of the strategies compared
in Section~\ref{sec:variants}.

\subsection{Auxiliary Instance Loss}
\label{sec:auxloss}
Plain attention pooling provides gradient signal only through the aggregated bag
embedding, which can weaken the per-patch representation. To counteract this, the
shared head is additionally applied to every patch embedding and the image label is
broadcast to all patches, giving the composite objective
\begin{equation}
\mathcal{L} = \mathcal{L}_{\text{bag}} + \lambda \, \mathcal{L}_{\text{instance}},
\qquad
\mathcal{L}_{\text{instance}} = \frac{1}{N}\sum_{i=1}^{N}
\ell\big(g(\mathbf{h}_i),\, y\big),
\end{equation}
in which $\ell$ denotes the cross-entropy loss and $\lambda$ controls the weight of
the instance term.

\subsection{Training Regime}
\label{sec:training}
Two training regimes are compared. In the \emph{two-stage} regime, the backbone and
shared head are first trained as a plain per-patch classifier (Stage 1), and the
resulting weights are then transferred to the attention model, which is fine-tuned
end-to-end with the auxiliary loss (Stage 2). In the \emph{single-stage} regime,
the attention model is trained directly at the image level (Stage 2 only) without
patch-level pre-training. The comparison of these regimes is itself one of the
ablations (Section~\ref{sec:res-regime}).

The models are optimised with AdamW~\cite{loshchilov2019adamw}; the remaining
optimisation details are given in Table~\ref{tab:hparams}. Hyperparameters were
selected by a single Optuna search~\cite{akiba2019optuna}, performed for the nano
backbone at a patch size of 224 pixels, and were transferred unchanged to all other
patch sizes and backbones, with the sole exception of the stochastic-depth
(drop-path) rate, which is increased with backbone capacity. Tuning is therefore not
repeated per configuration; this keeps the comparison controlled, at the cost of
accuracies that are, if anything, conservative for the non-nominal settings.

\begin{table}[H]
\centering
\caption{Training configuration and tuned hyperparameters, obtained for the nano
backbone at 224\,px and transferred as described in the text.}
\label{tab:hparams}
\small
\begin{tabular}{lcc}
\toprule
\textbf{Hyperparameter} & \textbf{Stage 1} & \textbf{Stage 2}\\
\midrule
Backbone            & \multicolumn{2}{c}{ConvNeXt-V2 (pico, nano, tiny)}\\
Base learning rate  & 1.0e-4 & 1.0e-5\\
Weight decay        & 1.0e-3 & 6.4e-4\\
Drop-path rate      & \multicolumn{2}{c}{0.15, 0.25, 0.35}\\
Instance-loss weight $\lambda$ & -- & 0.65\\
Differential-LR multiplier     & -- & 15\\
Attention Head Dropout         & -- & 0.42\\
Classifier Head Dropout        & 0.12 & 0.21\\
Batch / accumulation           & 128 & /32\\
\bottomrule
\end{tabular}
\end{table}

\subsection{Aggregation Strategies and Baselines}
\label{sec:variants}
Six ways of arriving at an image-level decision are compared over identical
backbones:
\begin{itemize}[leftmargin=*]
\item \textbf{Majority voting}: a hard majority vote over the per-patch predictions
of a bag, reproducing the aggregation used in earlier bark-classification work and
serving as the reference baseline.
\item \textbf{Attention-based MIL (\amil{})}: learned attention pooling with the
auxiliary loss (Sections~\ref{sec:model}--\ref{sec:auxloss}).
\item \textbf{Fixed-head majority vote} (\amil{}-vote): a majority vote taken over the
per-patch head of a trained \amil{} model, with the weights held fixed, which
isolates the effect of the aggregation rule from the additional training.
\item \textbf{Soft voting}: the argmax of the mean per-patch class probabilities, computed
on the Stage-1 classifier and on the per-patch head of a trained \amil{} model; this is
the soft counterpart of majority voting.
\item \textbf{Uniform attention}: a trained \amil{} model evaluated with all attention
weights fixed to $1/N$, so that the bag embedding is the mean of the patch embeddings;
comparing it with \amil{} isolates the contribution of the learned weights from that of
pooling soft evidence.
\item \textbf{Whole-image}: the entire photograph, resized to the network input, is
classified as a single instance; this is the limiting case in which a bag contains a
single patch, and it tests whether patch-based processing is beneficial at all.
\end{itemize}
Soft voting and uniform attention are evaluation-only: they reuse trained weights
without any further training.

\subsection{Ablation Design}
\label{sec:ablation-design}
The study is organised around a set of hypotheses, each isolating one design choice;
stating them explicitly delimits the scope of the study and allows each to be
confirmed or refuted individually. The corresponding results are reported in
Section~\ref{sec:results}.
\begin{itemize}[leftmargin=*]
\item \htag{H1} Per-patch accuracy increases with patch size, as larger patches
carry more discriminative bark structure.
\item \htag{H2} The advantage of learned aggregation over voting is greatest when
patches are individually weak and diminishes as patch size, and hence per-patch
quality, increases; equivalently, the attention distribution is more concentrated at
small patch sizes and more diffuse (voting-like) at large ones.
\item \htag{H3} Two-stage training (patch-level pre-training followed by attention
fine-tuning) yields a stronger patch representation and thereby higher final
accuracy than single-stage (image-level only) training.
\item \htag{H4} An optimal patch size exists that balances per-patch quality against
retaining enough patches to aggregate, so that the best size is a deliberate
trade-off rather than the largest possible.
\item \htag{H5} Patch-based processing outperforms whole-image classification.
\item \htag{H6} With the backbone held fixed, learned aggregation outperforms hard
voting.
\item \htag{H7} Accuracy increases with the number of patches available at
inference, with diminishing returns beyond a certain bag size.
\item \htag{H8} A larger backbone (more parameters) yields higher image-level
accuracy.
\item \htag{H9} An image-level split of the same data yields higher accuracy than the
tree-level split, overstating performance on unseen trees.
\end{itemize}

\subsection{Evaluation Metrics and Attention Diagnostics}
\label{sec:eval}
Every configuration is evaluated by five-fold, tree-level cross-validation, and all
quantities are reported as means and standard deviations across folds. Two
performance metrics are reported: top-1 image-level accuracy and the macro-averaged
$F_1$ score, the latter to account for class imbalance across the \nclasses{}
species.

To characterise how the attention model uses its patches, the Shannon entropy of the
attention distribution $\{\alpha_i\}$ of each bag is computed,
$H(\alpha) = -\sum_{i=1}^{N}\alpha_i \ln \alpha_i$, and summarised as the effective
fraction of patches attended, $\exp\!\big(H(\alpha)\big)/N \in (0,1]$. A value near
one indicates near-uniform attention, under which learned aggregation approaches an
average over all patches and is expected to behave like voting; small values
indicate that the decision is concentrated on a few patches. This diagnostic is used
to interpret the interaction between patch size and aggregation
(Section~\ref{sec:res-interaction}) and the effect of the training regime
(Section~\ref{sec:res-regime}).

\section{Results}
\label{sec:results}

\subsection{Effect of Patch Size on Per-Patch Accuracy \htag{H1}}
\label{sec:res-patchsize}
Per-patch (Stage-1) accuracy increases monotonically with patch size, from
$75.2\%$ at $160$\,px to $89.3\%$ at $512$\,px (Figure~\ref{fig:patchsize}), confirming
\textbf{H1}: larger patches, which capture more of the trunk's bark structure, are
individually more discriminative. The gains are largest at the small end of the
range (roughly $+5.6$\,pp from $160$ to $224$\,px) and taper towards the largest
sizes ($+2.0$\,pp from $384$ to $512$\,px).

\subsection{Interaction Between Patch Size and Aggregation \htag{H2}}
\label{sec:res-interaction}
\begin{figure}[H]
\centering
\includegraphics[width=\linewidth]{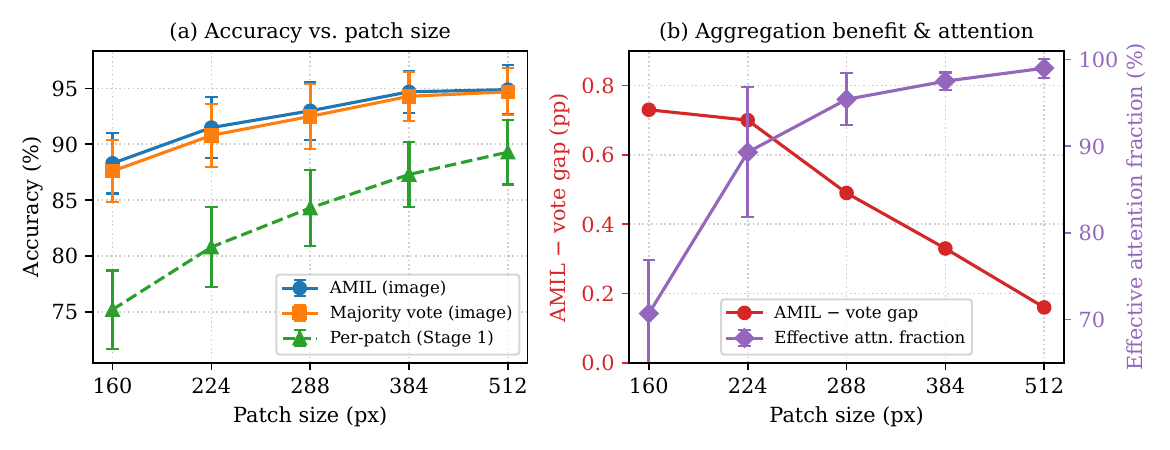}
\caption{Effect of patch size (nano backbone; five-fold, mean$\pm$std).
\textbf{(a)} Per-patch (Stage-1) accuracy and image-level accuracy under majority
voting and \amil{}: all rise with patch size and the image-level curves plateau near
$512$\,px, well above the per-patch accuracy that aggregation is built from.
\textbf{(b)} The \amil{}$-$voting gap (left axis) shrinks as patch size grows, while
the effective fraction of patches attended (right axis) rises toward one, so at large
patches the attention distribution is near-uniform and pooling behaves like voting.
\htag{H1}, \htag{H2}, \htag{H6}}
\label{fig:patchsize}
\end{figure}

The advantage of learned aggregation over majority voting is small at every patch
size and declines monotonically as patch size increases, from $+0.73$\,pp at
$160$\,px to $+0.16$\,pp at $512$\,px (Figure~\ref{fig:patchsize}). This decline tracks the
attention distribution directly: the effective fraction of patches attended rises
from $70.7\%$ at $160$\,px to $99.0\%$ at $512$\,px, so that at large patch sizes the
attention model spreads its weight almost uniformly and therefore behaves like
voting. Both observations support \textbf{H2}: learned aggregation is most useful
when instances are individually weak, and its value dissipates as per-patch quality
rises and the model has little reason to prefer some patches over others. The
concurrent fall in bag size, from a mean of $347$ patches at $160$\,px to $38$ at
$512$\,px, is noted here and disentangled from patch size in
Section~\ref{sec:decouple}.

\subsection{Training Regime: Single- versus Two-Stage \htag{H3}}
\label{sec:res-regime}
Two-stage training does not offer an advantage over single-stage training, contrary
to \textbf{H3}. The single-stage model attains $92.8\%$ image-level accuracy against
$91.5\%$ for the two-stage model (Table~\ref{tab:regime}), a difference that, if
anything, favours the simpler single-stage regime and is not robust across folds.
The two regimes differ sharply, however, in how they use attention. Under
single-stage training the attention distribution is far more concentrated (effective
fraction $50.6\%$ versus $89.3\%$), and pooling beats a vote over the same head by
$+1.63$\,pp rather than $+0.45$\,pp; against uniform attention, the gaps are
$+1.46$\,pp (positive in all five folds) and $+0.48$\,pp respectively. This is consistent with the mechanism identified
in Section~\ref{sec:res-interaction}: two-stage pre-training yields a stronger
per-patch representation, which leaves the attention module less to do, whereas
single-stage training, lacking that representation, leans more heavily on selective
attention to reach a comparable result. Two-stage training therefore buys a more
robust patch representation and more diffuse attention, but not higher accuracy, so
the added complexity of a separate pre-training stage is not warranted here.

\begin{table}[H]
\centering
\caption{Single- versus two-stage training at 224\,px, nano backbone (five-fold,
mean$\pm$std). The aggregation gaps are \amil{} minus a vote over the same fixed head,
and \amil{} minus the same model with uniform attention weights.
\htag{H3}}
\label{tab:regime}
\footnotesize
\setlength{\tabcolsep}{4pt}
\begin{tabular}{lccccc}
\toprule
\textbf{Regime} & \textbf{\amil{} acc.\ (\%)} & \textbf{\amil{} macro-$F_1$}
 & \textbf{$\Delta$(\amil{}$-$\amil{}-vote)\pp} & \textbf{$\Delta$(\amil{}$-$uniform)\pp} & \textbf{Eff.\ frac.\ (\%)}\\
\midrule
Two-stage    & 91.5$\pm$2.7 & 91.7 & $+0.45$ & $+0.48$ & 89.3\\
Single-stage & 92.8$\pm$2.2 & 92.7 & $+1.63$ & $+1.46$ & 50.6\\
\bottomrule
\end{tabular}
\end{table}

\subsection{Patch Size, Bag Size, and the Optimal Operating Point \htag{H4}}
\label{sec:res-sweetspot}
Per-patch accuracy increases monotonically with patch size
(Section~\ref{sec:res-patchsize}), and the whole-image baseline (the limit of a
single, maximally large ``patch'') attains the highest single-instance accuracy of
all ($91.9\%$; Section~\ref{sec:res-wholeimage}), exceeding even the $512$\,px
per-patch accuracy ($89.3\%$). Image-level accuracy, however, behaves differently. It
rises with patch size and plateaus near $512$\,px ($94.7\%$ at $384$ to $94.9\%$ at
$512$; Figure~\ref{fig:patchsize}), whereas the whole-image case, which admits no aggregation
because its bag contains a single instance, falls well below it ($91.9\%$). Enlarging
patches thus improves the individual instances but shrinks the bag, and past a certain
point too few patches remain to average over and image-level performance degrades, the
whole-image case being the extreme of this trade-off. The resulting optimum, where
per-patch quality is high yet enough patches remain to aggregate, lies around
$512$\,px, where image-level accuracy plateaus. This supports \textbf{H4}: the best
patch size is a deliberate trade-off rather than simply the largest possible.

\subsection{Patch-Based Processing versus Whole-Image Classification \htag{H5}}
\label{sec:res-wholeimage}
The whole-image baseline, in which the entire photograph is resized to the network
input and classified as a single instance, attains $91.9\% \pm 2.7$ accuracy
($91.9\%$ macro-$F_1$). Two observations follow. First, this exceeds the per-patch
accuracy of even the largest ($512$\,px) patches ($89.3\%$; Figure~\ref{fig:patchsize}): a
single view covering the whole trunk is more informative than a single localised
patch, confirming that a larger field of view per instance helps. Second, it
nonetheless trails patch-based aggregation at the larger patch sizes ($94.9\%$ at
$512$\,px; Figure~\ref{fig:patchsize}), so aggregating several large patches beats classifying
one whole image, supporting \textbf{H5}. One caveat applies: the whole-image baseline
resizes each photograph to the fixed $224$\,px network input, discarding most of the
native resolution and blurring fine bark texture, so its lower accuracy is partly an
artefact of that downsampling rather than evidence that whole-image classification is
inherently inferior. Indeed, this limitation is itself an argument for patching, the
mechanism by which a fixed-input network can exploit the full image resolution. The
comparison is also relevant to the leakage discussion: the leakage-safe whole-image
accuracy ($91.9\%$) is far below the near-saturated single-image figure of Cui et
al.~\cite{cui2023convnext} ($97.61\%$, reported on the 33-species BarkNetV2, which
includes a non-public subset, under an apparently image-level split); although the
datasets differ, the gap is consistent with same-trunk leakage inflating such a
split. Section~\ref{sec:res-leakage} tests this directly: under an image-level split of
\bnet{}, the same whole-image model reaches $99.2\%$, above the figure of Cui et al.

\subsection{Learned Aggregation versus Hard Voting \htag{H6}}
\label{sec:res-aggregation}
Holding the backbone fixed, attention pooling outperforms a majority vote taken over
the same per-patch head (the \amil{}-vote baseline) at every patch size, by
$+0.85$\,pp at $160$\,px falling to $+0.10$\,pp at $512$\,px; the comparison against
the classical Stage-1 voting baseline is similar in direction and magnitude ($+0.73$
to $+0.16$\,pp; Figure~\ref{fig:patchsize}). The differences are positive in the mean at every
patch size but small in magnitude, and their consistency across folds fades with patch
size: against Stage-1 soft voting, \amil{} wins in all five folds from $160$ to
$288$\,px, in four of five at $384$\,px, and in only two of five at $512$\,px.
\textbf{H6} is therefore supported in direction but not in magnitude: learned
aggregation confers a small edge that is reliable when patches are weak and
disappears as they strengthen.

The evaluation-only baselines separate the two ingredients of that edge
(Table~\ref{tab:h2}). Pooling soft evidence contributes nothing: soft voting matches
hard voting to within $0.1$\,pp at every patch size, whether computed on the Stage-1
classifier or on the \amil{} patch head, and uniform attention over the trained \amil{}
model matches the \amil{}-vote baseline to within $0.1$\,pp. The edge of \amil{} is
therefore due to the learned weights themselves. Its gap over uniform attention falls
from $+0.88$\,pp at $160$\,px (positive in all five folds) to $+0.02$\,pp at $512$\,px
(mixed in sign), tracking the effective attention fraction, which rises from $70.7\%$
to $99.0\%$. The single-stage model, whose attention is the most concentrated
(effective fraction $50.6\%$), shows the largest gap, $+1.46$\,pp in all five folds
(Table~\ref{tab:regime}), and the gap is similar across backbones at $224$\,px
($+0.38$, $+0.48$ and $+0.38$\,pp for pico, nano and tiny). Where attention departs
from uniform, the learned weights do measurable work; at $512$\,px they converge to
uniform weights and \amil{} reduces in practice to mean pooling.

This behaviour has a structural explanation. Attention-based MIL was devised for
settings such as computational pathology, in which a bag mixes a few informative
instances with a large majority of uninformative background patches, and the
attention layer earns its value by suppressing that
background~\cite{ilse2018attention,campanella2019clinical}. Bark bags are the opposite
case: the photographs are framed tightly on the trunk, so nearly every patch is a
discriminative instance, and near-equal weighting is close to optimal, precisely the
uniform attention observed here, and mean pooling of patch embeddings is known to be a
strong baseline in such settings~\cite{wang2018revisiting}. Whether learned aggregation is worthwhile on a
fine-grained task in which essentially all instances are informative is one of the
questions this study set out to answer; the answer is that its benefit is real but
marginal. Because the attention module adds only a handful of parameters over a plain
per-patch classifier and is never worse than voting in the mean, it is a low-cost
option worth enabling wherever maximal accuracy is valuable while it can be omitted with negligible loss when
simplicity is the priority.

\subsection{Effect of Test-Time Bag Size \htag{H7}}
\label{sec:res-bagsize}
\begin{figure}[H]
\centering
\includegraphics[width=0.75\linewidth]{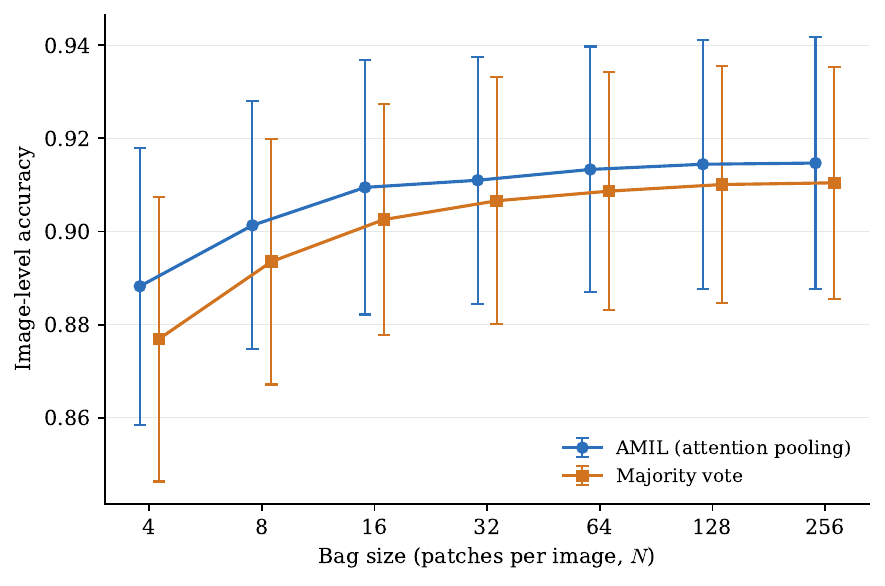}
\caption{Image-level accuracy as a function of the number of patches available at
inference (nano backbone, $224$\,px; mean$\pm$std over five folds). Both aggregation
rules improve steeply up to about $16$--$32$ patches and plateau thereafter, so a few
dozen patches per trunk suffice; \amil{} stays above majority voting throughout, by a
margin that is largest when patches are scarce and narrows as the bag grows. \htag{H7}}
\label{fig:h7}
\end{figure}

Image-level accuracy increases with the number of patches available at inference and
saturates well before the full bag is used (Figure~\ref{fig:h7}). For the nano $224$\,px model, \amil{} accuracy rises from
$88.8\%$ with $4$ patches to $91.3\%$ with $64$, and only $91.5\%$ with $256$; the
bulk of the benefit is realised by roughly $64$ patches, and a dozen already come
within half a point of the plateau. The advantage of learned aggregation over voting
is largest when few patches are available ($+1.13$\,pp at $4$ patches) and shrinks as
the bag grows ($+0.42$\,pp at $256$), echoing Section~\ref{sec:res-interaction}: when
evidence is scarce, weighting patches selectively matters more. In deployment, only a
few dozen patches per trunk are therefore needed to approach full accuracy, which
combines favourably with the preference for large patches (few patches per image)
established in Section~\ref{sec:res-sweetspot}. Because this sweep varies bag size at
a fixed $224$\,px patch size, it also serves to disentangle the two factors
(Section~\ref{sec:decouple}), confirming \textbf{H7}.

\subsection{Effect of Backbone Capacity \htag{H8}}
\label{sec:res-backbone}
Contrary to \textbf{H8}, backbone capacity has little effect on image-level accuracy.
Across ConvNeXt-V2 pico, nano, and tiny, a roughly threefold range in parameters,
\amil{} accuracy varies by less than a percentage point ($90.8\%$ to $91.5\%$), with
the nano backbone marginally ahead of the larger tiny (Table~\ref{tab:backbone}).
Per-patch accuracy rises slightly with capacity ($79.7\%$ to $81.2\%$), but this does
not translate into a meaningful image-level gain. A larger backbone is therefore not
worth its additional training and inference cost for this task, and the nano backbone
is retained as the default.

\begin{table}[H]
\centering
\caption{Effect of backbone capacity at a patch size of 224\,px (five-fold,
mean$\pm$std). \htag{H8}}
\label{tab:backbone}
\footnotesize
\setlength{\tabcolsep}{4pt}
\begin{tabular}{lcccccc}
\toprule
\textbf{Backbone} & \textbf{Params (M)} & \textbf{Per-patch (\%)}
 & \textbf{Voting (\%)} & \textbf{\amil{} (\%)} & \textbf{\amil{} macro-$F_1$}
 & \textbf{$\Delta$\pp}\\
\midrule
ConvNeXt-V2 pico & 9.1  & 79.7$\pm$3.2 & 90.1$\pm$3.0 & 90.8$\pm$2.9 & 90.9 & $+0.73$\\
ConvNeXt-V2 nano & 15.6 & 80.8$\pm$3.6 & 90.8$\pm$2.8 & 91.5$\pm$2.7 & 91.7 & $+0.70$\\
ConvNeXt-V2 tiny & 28.6 & 81.2$\pm$3.3 & 91.1$\pm$2.9 & 91.3$\pm$2.8 & 91.2 & $+0.19$\\
\bottomrule
\end{tabular}
\end{table}

\subsection{Effect of the Partition: Tree- versus Image-Level Splits \htag{H9}}
\label{sec:res-leakage}
Retraining the same models under the image-level comparison split
(Section~\ref{sec:split}) raises every accuracy (Table~\ref{tab:leakage}). The
whole-image baseline rises from $91.9\%$ to $99.2\%$ ($+7.3$\,pp), and the
recommended configuration from $94.9\%$ to $99.9\%$ ($+5.0$\,pp) at the image level,
with the per-patch accuracy rising from $89.3\%$ to $99.1\%$ ($+9.8$\,pp). These
differences dwarf the fold-to-fold variation of either protocol (standard deviations
of $2$--$3$\,pp under the tree split and at most $0.3$\,pp under the image split).

Three features of the result matter for the rest of the study. First, the inflation is
as large as the most important design effect measured here: moving from $160$ to
$512$\,px patches raises \amil{} accuracy by $6.6$\,pp, whereas the partition alone
moves it by $5.0$\,pp, and the whole-image figure by more than that; it is several
times larger than the effect of learned aggregation (at most $1.6$\,pp, in the
single-stage model) or of backbone
capacity (under $1$\,pp). Second, single instances are inflated most (patches
$+9.8$\,pp, whole images $+7.3$\,pp), because a near-duplicate view of the same trunk
can be recognised from a single patch, while aggregated accuracy is already pressed
against the ceiling. Third, that ceiling erases the comparisons the ablation is built
on: under the image split, voting and \amil{} both reach $99.9\%$, and their per-fold
differences are within $\pm0.04$\,pp, so the design choices that are distinguishable
under the tree split become indistinguishable. The attention distribution is almost
unchanged (effective fraction $99.7\%$ against $99.0\%$). Finally, the image-split
whole-image figure exceeds the $97.61\%$ reported by Cui et al.~\cite{cui2023convnext};
although their dataset is different and larger, their result lies within the range
that an image-level split produces. \textbf{H9} is therefore supported.

\begin{table}[H]
\centering
\caption{The same models under the tree-level split used throughout and under an
image-level split of the same data (nano backbone; five-fold, mean$\pm$std). $\Delta$ is
image-level minus tree-level. \htag{H9}}
\label{tab:leakage}
\small
\begin{threeparttable}
\begin{tabular}{lccc}
\toprule
\textbf{Quantity} & \textbf{Tree split} & \textbf{Image split} & \textbf{$\Delta$\pp}\\
\midrule
Whole-image accuracy (\%)                   & 91.9$\pm$2.7 & 99.2$\pm$0.2 & $+7.3$\\
Whole-image macro-$F_1$ (\%)                & 91.9$\pm$2.6 & 99.2$\pm$0.2 & $+7.3$\\
Per-patch accuracy, $512$\,px (\%)          & 89.3$\pm$2.9 & 99.1$\pm$0.1 & $+9.8$\\
Voting accuracy, $512$\,px (\%)             & 94.7$\pm$2.1 & 99.9$\pm$0.0 & $+5.1$\\
\amil{} accuracy, $512$\,px (\%)            & 94.9$\pm$2.2 & 99.9$\pm$0.0 & $+5.0$\\
\amil{} macro-$F_1$, $512$\,px (\%)         & 95.0$\pm$2.0 & 99.9$\pm$0.1 & $+4.8$\\
Effective attention fraction (\%)           & 99.0$\pm$1.1 & 99.7$\pm$0.3 & $+0.7$\\
\bottomrule
\end{tabular}
\begin{tablenotes}
\item Tree-level aggregation is not reported for the image split: the images of a tree
are spread across folds, so a per-tree decision would not be comparable.
\end{tablenotes}
\end{threeparttable}
\end{table}

\subsection{Interpretability: Attention over the Trunk}
\label{sec:interp}
\begin{figure}[H]
\centering
\includegraphics[width=\linewidth]{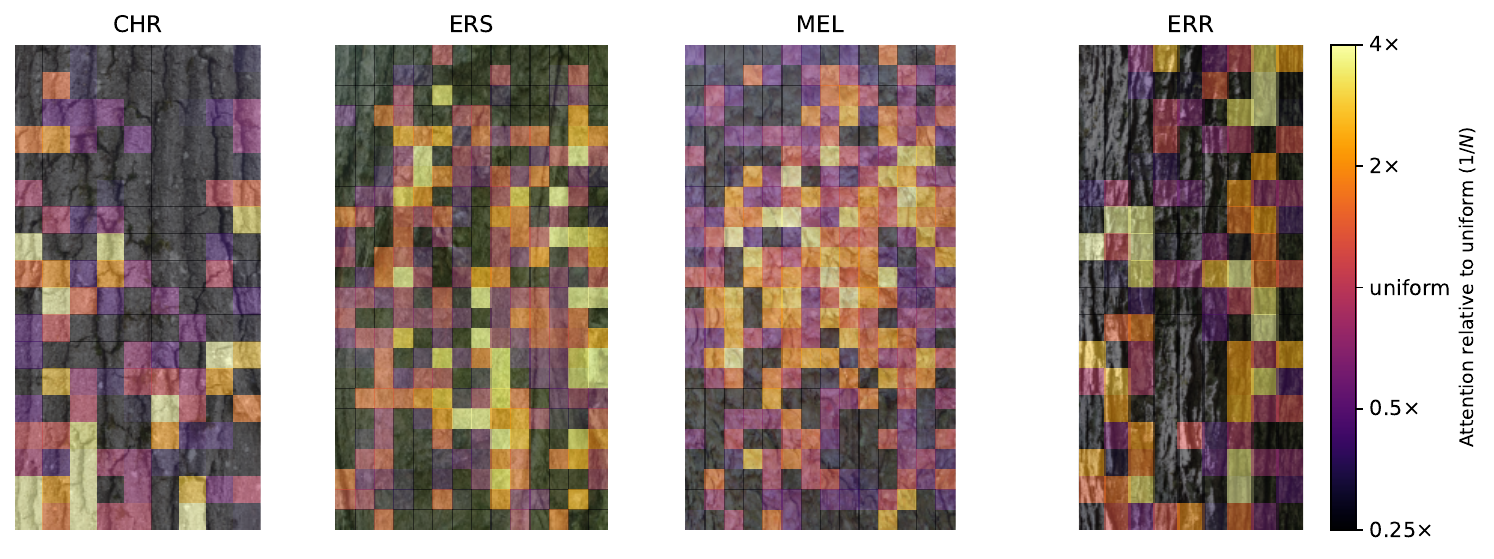}
\caption{Attention over the trunk for four species (CHR, ERS, MEL, ERR), nano
backbone at 224\,px resulting from single-stage training. Each patch is shaded by its
attention weight relative to the uniform weight $1/N$ (colour bar, $0.25\times$ to
$4\times$); brighter patches receive more weight. The attention is largely diffuse:
most patches receive near-uniform weight and the few higher-weight patches show no
consistent spatial structure.}
\label{fig:attention}
\end{figure}

Figure~\ref{fig:attention} shows the learned attention over representative trunks of
four species, expressed relative to the uniform weight $1/N$. In practice, the
attention offers little interpretive value. At the small patch size used here the
weights are distributed fairly evenly across the trunk: most patches receive
near-uniform weight, and the few higher-weight patches form no consistent,
human-legible pattern that could be tied to particular bark features. This is the
spatial counterpart of the diffuse attention quantified in
Section~\ref{sec:res-interaction} and, we suggest, a consequence of the patch-based
design itself. Because each patch is classified independently, the model has no
access to the larger-scale bark structure (ridges, plating, lenticel rows) that a
human observer uses, so attention merely re-weights already-similar local views.
Enlarging the patches does not recover interpretability either: as patch size grows
the attention becomes essentially uniform (Figure~\ref{fig:patchsize}), conveying even less
about which regions matter. Viewing the attention map is therefore of limited
diagnostic use here, consistent with evidence that deep MIL pooling often behaves as a
learned weighted average rather than as an instance selector~\cite{raff2023mil}, in contrast to whole-image saliency methods such as
class-activation mapping, which have been used to localise discriminative bark
regions on entire trunk images~\cite{barkcam2022}.

\subsection{Per-Species Performance}
\label{sec:res-perspecies}
\begin{figure}[H]
\centering
\includegraphics[width=0.9\linewidth]{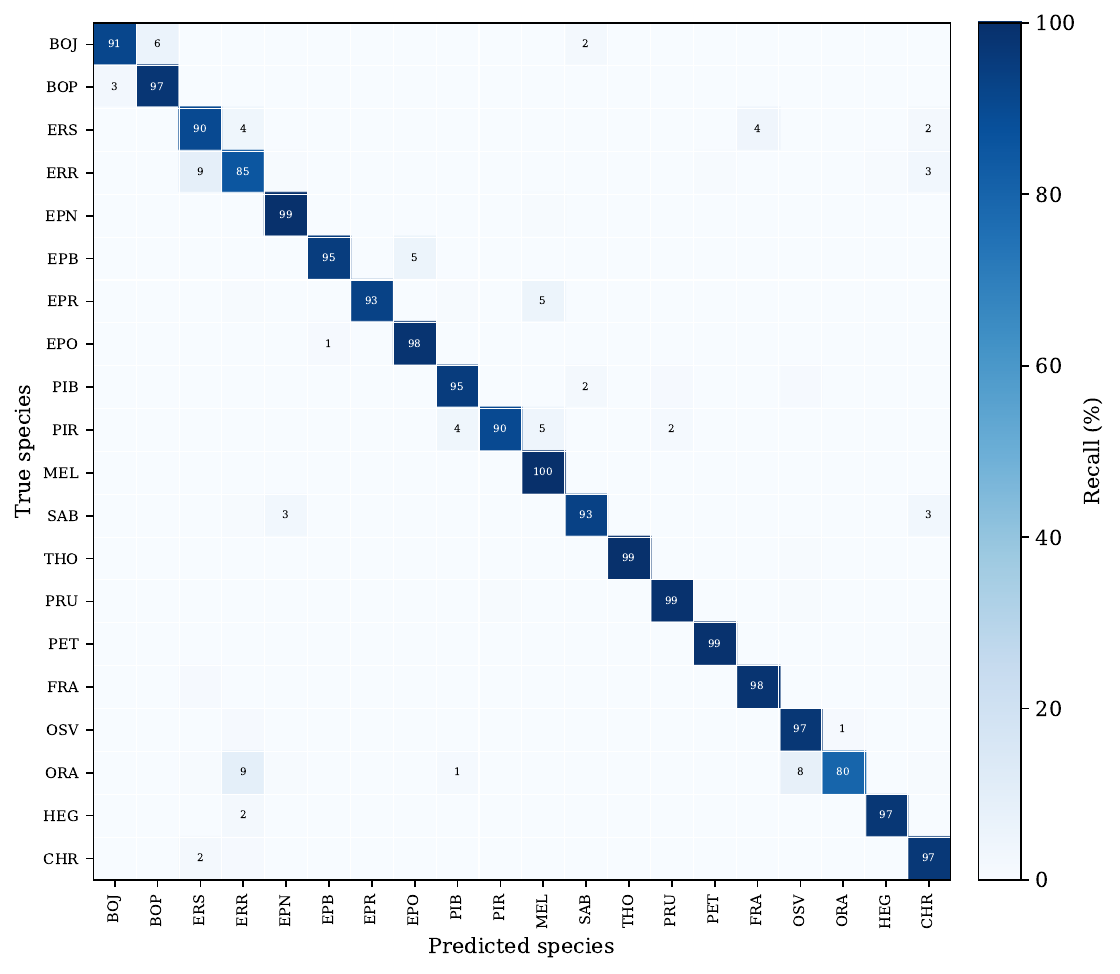}
\caption{Row-normalized (recall, \%) confusion matrix for the recommended
configuration (nano, $512$\,px, \amil{}, image level), pooled over the five held-out
test folds (each image classified once). Rows are true species, columns predicted;
species are ordered by genus, so that the predominantly intra-genus confusions appear
as near-diagonal blocks.}
\label{fig:confusion}
\end{figure}

For the recommended configuration (nano backbone, $512$\,px, \amil{}), the
image-level confusion matrix is shown in Figure~\ref{fig:confusion}; per-class
precision, recall, and $F_1$ are reported in Table~\ref{tab:perclass}. Classification
is strong across the board ($94.8\%$ accuracy, $94.9\%$ macro-$F_1$), and the residual
errors are highly structured: almost all confusions occur \emph{within} a genus rather
than across genera. Yellow and white birch (BOJ/BOP), sugar and red maple (ERS/ERR),
the spruces (EPB/EPO), and red pine with tamarack (PIR/MEL) account for the bulk of the
off-diagonal mass, whereas confusions between genera are rare.

This pattern is consistent with the visual structure of the task. Congeneric species
tend to share their gross bark characters (ridge and plate structure, colour, and
lenticel arrangement), and the features that separate them are often subtle and
expressed at a spatial scale, or under lighting conditions, that the present patch size
and backbone may not fully capture; in some cases the more reliable distinguishing
cues are not morphological at all but chemical, as documented for spruces via terpene
profiles~\cite{malhocka2025spruce}. We stress that these confusions reflect the limits
of \emph{this model at this resolution} rather than a claim that the species are
visually indistinguishable: trained observers routinely separate these taxa from bark.
Nonetheless, the errors are concentrated exactly where congeneric bark is most alike,
as the near-diagonal, within-genus structure of Figure~\ref{fig:confusion} makes plain.
The weakest class, American elm (ORA, $79.7\%$ recall), follows the same pattern,
being confused mainly with red maple and ironwood, whose bark is similarly ridged.
These congeneric confusions account for essentially all of the small gap between
overall accuracy and macro-$F_1$.

\section{Discussion}
\label{sec:discussion}

\subsection{Recommended Configuration}
\label{sec:recommendation}
Taken together, the ablations point to a simple recommended configuration. Patch size
is the dominant factor and should be made as large as the image allows while retaining
a bag large enough to average over; on \bnet{} this optimum is around $512$\,px
(Section~\ref{sec:res-sweetspot}). The backbone need not be large: ConvNeXt-V2 nano
offers the best accuracy-to-size trade-off, with larger capacities giving no
meaningful gain (Section~\ref{sec:res-backbone}). The aggregation rule and the
training regime are second-order. Single-stage training suffices, since two-stage
pre-training offers no advantage (Section~\ref{sec:res-regime}); and while majority
voting is already strong, attention pooling is at least as accurate in the mean for a
negligible increase in parameters. Its gain is reliable only where attention is
concentrated, with small patches or single-stage training; at the recommended
$512$\,px it reduces to mean pooling, and voting is an equally good choice
(Section~\ref{sec:res-aggregation}). Finally, only about a dozen patches per
trunk are needed at inference to approach full accuracy, with essentially no further
gain beyond roughly $64$ (Section~\ref{sec:res-bagsize}).

\subsection{Comparison with Prior Work}
\label{sec:priorwork}
The strongest results obtained across the ablation are compared with prior work in
Table~\ref{tab:priorwork}. Under the leakage-safe, tree-level protocol, they closely
reproduce the honest, tree-split figures of Carpentier et
al.~\cite{carpentier2018barknet} at every level of aggregation: single-patch accuracy
of $89.3\%$ against their $87.0\%$ single crop, image-level voting of $94.7\%$ against
their $93.88\%$, and tree-level \amil{} of $97.8\%$ against their $97.81\%$. The only
other tree-level deep-learning result on \bnet{} that we are aware of, the $94.62\%$
multi-crop accuracy of Wu et al.~\cite{wu2021deepbarkid}, likewise agrees with our
image-level voting figure. The
near-saturated single-image figure of Cui et al.~\cite{cui2023convnext} is not
directly comparable: it is reported on the 33-species BarkNetV2 (which includes a
non-public subset) under an apparently image-level split, and the whole-image result
of Section~\ref{sec:res-wholeimage} ($91.9\%$ under a leakage-safe split) is
consistent with same-trunk leakage inflating such a split. Under our own image-level
split of \bnet{}, the whole-image model reaches $99.2\%$ and \amil{} $99.9\%$, both above
the figure of Cui et al.\ (Section~\ref{sec:res-leakage}).

\begin{table}[H]
\centering
\caption{Comparison with prior work on \bnet{} (\nclasses{} species where
comparable). The results of Carpentier et al.\ were obtained under a tree-level
split and are directly comparable; the split of Cui et al.\ appears to be
image-level and may therefore reflect same-trunk leakage. Rows marked ``This work''
use the nano backbone at $512$\,px and are five-fold means.}
\label{tab:priorwork}
\small
\begin{threeparttable}
\begin{tabular}{llll}
\toprule
\textbf{Method} & \textbf{Split} & \textbf{Aggregation} & \textbf{Acc.\ (\%)}\\
\midrule
Carpentier et al.~\cite{carpentier2018barknet}  & tree & single crop  & 87.0\\
Carpentier et al.~\cite{carpentier2018barknet}  & tree & vote (image) & 93.88\\
Carpentier et al.~\cite{carpentier2018barknet}  & tree & vote (tree)  & 97.81\\
Wu et al.~\cite{wu2021deepbarkid}\tnote{b}      & tree & vote (image) & 94.62\\
Cui et al.~\cite{cui2023convnext}\tnote{a}      & image\tnote{a} & single image & 97.61\\
\midrule
\textbf{This work: per-patch} & tree & single patch & 89.3\\
\textbf{This work: image}     & tree & vote         & 94.7\\
\textbf{This work: image}     & tree & \amil{}      & 94.9\\
\textbf{This work: tree}      & tree & \amil{}      & 97.8\\
\midrule
\textit{This work: whole image} & image\tnote{c} & single image & 99.2\\
\textit{This work: image}       & image\tnote{c} & \amil{}      & 99.9\\
\bottomrule
\end{tabular}
\begin{tablenotes}
\item[a] BarkNetV2 (33 classes, including a non-public subset); the partition appears
to be random by image, so the figure is not directly comparable and may reflect
same-trunk leakage.
\item[b] ResNet-34, multi-crop; the 20-species subset reported there contains 23{,}359
images, against 23{,}271 here.
\item[c] Image-level comparison split (Section~\ref{sec:res-leakage}), shown only to
quantify leakage; not a recommended evaluation protocol.
\end{tablenotes}
\end{threeparttable}
\end{table}

\subsection{Decoupling the Bag-Size and Patch-Size Confound}
\label{sec:decouple}
Patch size and bag size are inevitably confounded in the patch-size sweep, since
larger patches produce fewer instances per image (Figure~\ref{fig:patchsize}). The test-time
bag-size sweep of Section~\ref{sec:res-bagsize} isolates the two: holding patch size
fixed at $224$\,px and reducing the number of patches lowers accuracy only modestly
and with strong diminishing returns ($91.5\%$ at $256$ patches to $88.8\%$ at $4$). By
contrast, the patch-size sweep shows accuracy \emph{rising} as patches grow
\emph{despite} the accompanying fall in bag size. Over the range tested, the two
effects act in opposite directions, and the net improvement with larger patches is
attributable to per-patch quality rather than to bag size. The whole-image case
(Section~\ref{sec:res-sweetspot}) marks the point at which the shrinking bag finally
dominates and accuracy falls, locating the optimum at an intermediate-to-large patch
size.

\subsection{Limitations}
\label{sec:limitations}
Several limitations qualify these findings. The study uses a single dataset from one
geographic region and species set, so the recommended configuration may not transfer
directly to other floras or acquisition conditions; bark moisture and lighting are known
to shift accuracy~\cite{warner2025environmental}, and even specimen-level cross-validation
can overstate accuracy on an independent collection~\cite{ravindran2022diffuseporous}. Classes are imbalanced, which
macro-$F_1$ mitigates but does not remove. Hyperparameters were tuned only for the
nano backbone at $224$\,px and transferred, which may modestly understate the accuracy
of the other settings. The training and evaluation bags differ in size (training caps
the number of patches per bag, whereas evaluation uses the full bag). Because images
from the same tree are correlated, we characterise the smaller effects, notably the
aggregation advantage, by their magnitude and their consistency in sign across folds
rather than by image-level significance tests, which would overstate confidence on
clustered data. Finally, we evaluate a single, established attention-MIL
formulation~\cite{ilse2018attention}; aggregators that model inter-patch context, such
as transformer-based or dual-stream MIL~\cite{shao2021transmil,li2021dsmil}, might behave differently, although
the near-uniform attention observed here, driven by the absence of uninformative
background patches, suggests limited headroom. We leave this to future work.

\section{Conclusions}
\label{sec:conclusions}
Under a strictly leakage-safe, tree-level evaluation on \bnet{}, this study
systematically ablated the principal design choices for bark-based tree-species
classification. Accuracy is governed chiefly by patch size: it should be made as large
as the image allows while still leaving enough patches to average over, an optimum
reached around $512$\,px, beyond which too few patches remain and performance
degrades, as the whole-image case makes plain. Backbone capacity and the training
regime are second-order: ConvNeXt-V2 nano gives the best accuracy-to-size trade-off,
and two-stage training offers no advantage over the simpler single-stage regime.
Learned attention pooling is at least as accurate as majority voting in the mean for
a negligible parameter cost, its gain coming from the learned weights rather than from
pooling soft evidence, making it worthwhile wherever maximal accuracy matters,
and only about a dozen patches per trunk are needed at inference to approach full
accuracy. Splitting the same data by image instead of by tree inflates image-level
accuracy by $5.0$\,pp ($7.3$\,pp for whole images), as much as the full effect of patch
size and several times more than aggregation or capacity, and pushes accuracy so close
to the ceiling that the design choices can no longer be told apart, so
tree-level partitioning should be the default for bark benchmarks. Because the task is an instance of tile-based fine-grained recognition,
these findings may also inform patch-and-aggregate pipelines beyond bark, and they
provide a dependable, inexpensive starting point for deployment and future research.

\section*{Author Contributions}
This study was conceived, implemented, analysed, and written by the sole author,
M.\ C\^ot\'e.

\section*{Funding}
This research received no external funding. The computational experiments were
carried out on infrastructure provided by the Digital Research Alliance of Canada
(DRAC).

\section*{Data and Code Availability}
The \bnet{} dataset is available at
\url{https://github.com/ulaval-damas/tree-bark-classification}. Code to reproduce
the experiments, including the per-image cross-validation fold assignments, is
available at \url{https://github.com/MaeveCote/barknet-amil}.

\section*{Acknowledgments}
The author thanks A.\ Oberman (McGill University and Mila) for sponsoring the
Digital Research Alliance of Canada resource allocation under which the experiments were run.

\section*{Conflicts of Interest}
The author declares no conflict of interest.

\section*{Declaration of Generative AI Use}
The author used Anthropic Claude, an AI-assisted editing tool, to improve spelling, grammar, clarity, and readability during manuscript preparation and to help format the \LaTeX{} manuscript. After using this tool, the author carefully reviewed and edited the text and takes full responsibility for the final content.

\appendix
\renewcommand{\thetable}{A\arabic{table}}   % tables become A1, A2, A3
\setcounter{table}{0}

\section{Detailed Result Tables}
\label{app:tables}
The following tables give the exact five-fold means and standard deviations behind
Figures~\ref{fig:patchsize}, ~\ref{fig:h7} and ~\ref{fig:confusion}.

\begin{table}[H]
\centering
\caption{Per-patch (Stage-1) accuracy by patch size, nano backbone (five-fold,
mean$\pm$std). Exact values for Figure~\ref{fig:patchsize}(a). \htag{H1}}
\label{tab:h1}
\small
\begin{tabular}{lccccc}
\toprule
\textbf{Patch size (px)} & \textbf{160} & \textbf{224} & \textbf{288} & \textbf{384} & \textbf{512}\\
\midrule
Per-patch accuracy (\%) & 75.2$\pm$3.5 & 80.8$\pm$3.6 & 84.3$\pm$3.4 & 87.3$\pm$2.9 & 89.3$\pm$2.9\\
\bottomrule
\end{tabular}
\end{table}

\begin{table}[H]
\centering
\caption{Image-level performance of all aggregation rules, aggregation comparison, and attention
concentration by patch size (nano backbone; five-fold, mean$\pm$std). Exact values
for Figure~\ref{fig:patchsize}. \htag{H2}, \htag{H6}}
\label{tab:h2}
\small
\begin{tabular}{lccccc}
\toprule
\textbf{Patch size (px)} & \textbf{160} & \textbf{224} & \textbf{288} & \textbf{384} & \textbf{512}\\
\midrule
Mean bag size (patches)            & 347 & 181 & 113 & 64 & 38\\
\midrule
Voting accuracy (\%)               & 87.6$\pm$2.8 & 90.8$\pm$2.8 & 92.5$\pm$2.9 & 94.3$\pm$2.2 & 94.7$\pm$2.1\\
\amil{}-vote accuracy (\%)         & 87.5$\pm$3.2 & 91.0$\pm$2.5 & 92.7$\pm$2.6 & 94.5$\pm$2.1 & 94.8$\pm$2.0\\
Soft voting, Stage 1 (\%)          & 87.6$\pm$2.9 & 90.8$\pm$2.8 & 92.5$\pm$2.9 & 94.3$\pm$2.1 & 94.8$\pm$2.1\\
Soft voting, \amil{} head (\%)     & 87.5$\pm$3.2 & 91.0$\pm$2.5 & 92.7$\pm$2.6 & 94.5$\pm$2.0 & 94.9$\pm$2.0\\
Uniform attention (\%)             & 87.5$\pm$3.2 & 91.0$\pm$2.5 & 92.7$\pm$2.7 & 94.5$\pm$2.0 & 94.9$\pm$2.0\\
\amil{} accuracy (\%)              & 88.3$\pm$2.7 & 91.5$\pm$2.7 & 93.0$\pm$2.6 & 94.7$\pm$1.9 & 94.9$\pm$2.2\\
$\Delta$ (\amil{}$-$voting)\pp     & $+0.73$ & $+0.70$ & $+0.49$ & $+0.33$ & $+0.16$\\
$\Delta$ (\amil{}$-$uniform)\pp    & $+0.88$ & $+0.48$ & $+0.26$ & $+0.14$ & $+0.02$\\
\midrule
Effective attention fraction (\%)  & 70.7$\pm$6.2 & 89.3$\pm$7.5 & 95.4$\pm$3.0 & 97.5$\pm$1.0 & 99.0$\pm$1.1\\
\bottomrule
\end{tabular}
\end{table}

\begin{table}[H]
\centering
\caption{Image-level accuracy versus the number of patches available at inference
(nano, $224$\,px; five-fold means). Exact values for Figure~\ref{fig:h7}. \htag{H7}}
\label{tab:bagsize}
\small
\begin{tabular}{lccccccc}
\toprule
\textbf{Patches per bag} & \textbf{4} & \textbf{8} & \textbf{16} & \textbf{32}
 & \textbf{64} & \textbf{128} & \textbf{256}\\
\midrule
Voting accuracy (\%)           & 87.5 & 89.0 & 90.0 & 90.4 & 90.6 & 90.7 & 90.8\\
\amil{} accuracy (\%)          & 88.8 & 90.1 & 91.0 & 91.1 & 91.3 & 91.4 & 91.5\\
$\Delta$ (\amil{}$-$voting)\pp & $+1.13$ & $+0.78$ & $+0.69$ & $+0.44$ & $+0.47$ & $+0.44$ & $+0.42$\\
\bottomrule
\end{tabular}
\end{table}

\begin{table}[H]
\centering
\caption{Per-class precision, recall, and $F_1$ (\%) for the recommended
configuration (nano, $512$\,px, \amil{}, image level). $N$ is the number of test
images per species (each classified once across the five held-out folds).}
\label{tab:perclass}
\small
\begin{tabular}{lrrrr}
\toprule
\textbf{Species} & \textbf{Precision} & \textbf{Recall} & \textbf{$F_1$} & \textbf{$N$}\\
\midrule
BOJ & 97.0 & 91.4 & 94.1 & 1255\\
BOP & 94.0 & 97.4 & 95.7 & 1285\\
ERS & 89.0 & 90.1 & 89.5 & 1911\\
ERR & 88.0 & 85.3 & 86.6 & 1676\\
EPN & 95.0 & 99.3 & 97.1 & 885\\
EPB & 96.9 & 94.6 & 95.8 & 596\\
EPR & 98.6 & 92.8 & 95.6 & 740\\
EPO & 97.2 & 98.4 & 97.8 & 1324\\
PIB & 95.5 & 95.3 & 95.4 & 1023\\
PIR & 98.7 & 89.9 & 94.1 & 596\\
MEL & 96.1 & 99.9 & 98.0 & 1874\\
SAB & 95.0 & 93.1 & 94.0 & 922\\
THO & 99.2 & 99.5 & 99.3 & 746\\
PRU & 97.6 & 98.9 & 98.2 & 986\\
PET & 97.9 & 99.0 & 98.5 & 1037\\
FRA & 94.1 & 98.1 & 96.1 & 1472\\
OSV & 87.9 & 97.1 & 92.2 & 612\\
ORA & 96.5 & 79.7 & 87.3 & 767\\
HEG & 98.1 & 97.3 & 97.7 & 840\\
CHR & 95.0 & 96.5 & 95.8 & 2724\\
\midrule
\textbf{Macro avg.} & \textbf{95.4} & \textbf{94.7} & \textbf{94.9} & \textbf{23{,}271}\\
\bottomrule
\end{tabular}
\end{table}

\bibliographystyle{unsrtnat}
\bibliography{bibliography}

@article{carpentier2018barknet,
  title   = {Tree Species Identification from Bark Images Using Convolutional Neural Networks},
  author  = {Carpentier, Mathieu and Gigu{\`e}re, Philippe and Gaudreault, Jonathan},
  journal = {arXiv preprint arXiv:1803.00949},
  year    = {2018},
  note    = {Also in IEEE/RSJ Int.\ Conf.\ on Intelligent Robots and Systems (IROS) 2018}
}

@article{cui2023convnext,
  title   = {Improvement and Assessment of Convolutional Neural Network for Tree
             Species Identification Based on Bark Characteristics},
  author  = {Zhelin Cui and Xinran Li and Tao Li and Mingyang Li},
  journal = {Forests},
  volume  = {14},
  number  = {7},
  pages   = {1292},
  year    = {2023},
  publisher = {MDPI},
  doi     = {10.3390/f14071292}
}

@article{malhocka2025spruce,
  title   = {Inter and intraspecies variability of terpene profiles of four spruce
             species growing in the Czech Republic: a chemotaxonomic study},
  author  = {Malhock\'a, Alexandra and \v{S}v\'abov\'a, Martina},
  journal = {Biochemical Systematics and Ecology},
  volume  = {123},
  pages   = {105095},
  year    = {2025},
  publisher = {Elsevier}
}

@inproceedings{ratajczak2019bark,
  title     = {Efficient Bark Recognition in the Wild},
  author    = {Ratajczak, R{\'e}mi and Bertrand, Sarah and Crispim-Junior, Carlos
               and Tougne, Laure},
  booktitle = {Proceedings of the 14th International Joint Conference on Computer
               Vision, Imaging and Computer Graphics Theory and Applications
               (VISIGRAPP) -- Volume 4: VISAPP},
  pages     = {240--248},
  year      = {2019},
  publisher = {SciTePress},
  doi       = {10.5220/0007361902400248}
}

@article{warner2024centralbark,
  title   = {CentralBark Image Dataset and Tree Species Classification Using
             Deep Learning},
  author  = {Warner, Charles and Wu, Fanyou and Gazo, Rado and Benes, Bedrich
             and Kong, Nicole and Fei, Songlin},
  journal = {Algorithms},
  volume  = {17},
  number  = {5},
  pages   = {179},
  year    = {2024},
  publisher = {MDPI}
}

@inproceedings{misra2020patch,
  title     = {Patch-Based CNN Evaluation for Bark Classification},
  author    = {Misra, Debaleena and Crispim-Junior, Carlos and Tougne, Laure},
  booktitle = {Computer Vision -- ECCV 2020 Workshops (Workshop on Computer Vision
               Problems in Plant Phenotyping)},
  series    = {Lecture Notes in Computer Science},
  volume    = {12540},
  pages     = {197--212},
  year      = {2020},
  publisher = {Springer},
  doi       = {10.1007/978-3-030-65414-6_15}
}

@inproceedings{shao2021transmil,
  title     = {TransMIL: Transformer based Correlated Multiple Instance Learning
               for Whole Slide Image Classification},
  author    = {Shao, Zhuchen and Bian, Hao and Chen, Yang and Wang, Yifeng
               and Zhang, Jian and Ji, Xiangyang and Zhang, Yongbing},
  booktitle = {Advances in Neural Information Processing Systems (NeurIPS)},
  volume    = {34},
  pages     = {2136--2147},
  year      = {2021}
}

@inproceedings{fiel2011bark,
  title     = {Automated identification of tree species from images of the bark,
               leaves or needles},
  author    = {Fiel, Stefan and Sablatnig, Robert},
  booktitle = {Proceedings of the 16th Computer Vision Winter Workshop (CVWW)},
  pages     = {67--74},
  year      = {2011}
}

@inproceedings{boudra2018srbp,
  title     = {Bark identification using improved statistical radial binary patterns},
  author    = {Boudra, Safia and Yahiaoui, Itheri and Behloul, Ali},
  booktitle = {International Conference on Content-Based Multimedia Indexing (CBMI)},
  pages     = {1--6},
  year      = {2018},
  organization = {IEEE}
}

@article{barkcam2022,
  title   = {Identifying and extracting bark key features of 42 tree species using
             convolutional neural networks and class activation mapping},
  author  = {Tae Kyung Kim and Jeonghyun Hong and Daun Ryu and Sukyung Kim
             and Si Yeon Byeon and Woojin Huh and Kunhyo Kim and Gyu Heon Baek
             and Hyun Seok Kim},
  journal = {Scientific Reports},
  volume  = {12},
  pages   = {4772},
  year    = {2022},
  publisher = {Nature Publishing Group},
  doi     = {10.1038/s41598-022-08571-9}
}

@article{wei2022fgia,
  title   = {Fine-Grained Image Analysis with Deep Learning: A Survey},
  author  = {Wei, Xiu-Shen and Song, Yi-Zhe and Mac Aodha, Oisin and Wu, Jianxin
             and Peng, Yuxin and Tang, Jinhui and Yang, Jian and Belongie, Serge},
  journal = {IEEE Transactions on Pattern Analysis and Machine Intelligence},
  volume  = {44},
  number  = {12},
  pages   = {8927--8948},
  year    = {2022},
  publisher = {IEEE}
}

@article{waldchen2018trends,
  title   = {Automated plant species identification---Trends and future directions},
  author  = {W{\"a}ldchen, Jana and Rzanny, Michael and Seeland, Marco and M{\"a}der, Patrick},
  journal = {PLoS Computational Biology},
  volume  = {14},
  number  = {4},
  pages   = {e1005993},
  year    = {2018},
  publisher = {Public Library of Science}
}

@inproceedings{ilse2018attention,
  title     = {Attention-based Deep Multiple Instance Learning},
  author    = {Ilse, Maximilian and Tomczak, Jakub M. and Welling, Max},
  booktitle = {Proceedings of the 35th International Conference on Machine Learning (ICML)},
  pages     = {2127--2136},
  year      = {2018}
}

@article{campanella2019clinical,
  title   = {Clinical-grade computational pathology using weakly supervised deep
             learning on whole slide images},
  author  = {Campanella, Gabriele and Hanna, Matthew G. and Geneslaw, Luke and
             Miraflor, Allen and Werneck Krauss Silva, Vitor and Busam, Klaus J.
             and Brogi, Edi and Reuter, Victor E. and Klimstra, David S.
             and Fuchs, Thomas J.},
  journal = {Nature Medicine},
  volume  = {25},
  number  = {8},
  pages   = {1301--1309},
  year    = {2019},
  publisher = {Nature Publishing Group}
}

@article{lu2021clam,
  title   = {Data-efficient and weakly supervised computational pathology on
             whole-slide images},
  author  = {Lu, Ming Y. and Williamson, Drew F.K. and Chen, Tiffany Y.
             and Chen, Richard J. and Barbieri, Matteo and Mahmood, Faisal},
  journal = {Nature Biomedical Engineering},
  volume  = {5},
  number  = {6},
  pages   = {555--570},
  year    = {2021},
  publisher = {Nature Publishing Group}
}

@article{kapoor2023leakage,
  title   = {Leakage and the reproducibility crisis in machine-learning-based science},
  author  = {Kapoor, Sayash and Narayanan, Arvind},
  journal = {Patterns},
  volume  = {4},
  number  = {9},
  pages   = {100804},
  year    = {2023},
  publisher = {Elsevier}
}

@inproceedings{woo2023convnextv2,
  title     = {ConvNeXt V2: Co-designing and Scaling ConvNets with Masked Autoencoders},
  author    = {Woo, Sanghyun and Debnath, Shoubhik and Hu, Ronghang and Chen, Xinlei
             and Liu, Zhuang and Kweon, In So and Xie, Saining},
  booktitle = {Proceedings of the IEEE/CVF Conference on Computer Vision and Pattern Recognition (CVPR)},
  pages     = {16133--16142},
  year      = {2023}
}

@inproceedings{liu2022convnet,
  title     = {A ConvNet for the 2020s},
  author    = {Liu, Zhuang and Mao, Hanzi and Wu, Chao-Yuan and Feichtenhofer, Christoph
             and Darrell, Trevor and Xie, Saining},
  booktitle = {Proceedings of the IEEE/CVF Conference on Computer Vision and Pattern Recognition (CVPR)},
  pages     = {11976--11986},
  year      = {2022}
}

@inproceedings{deng2009imagenet,
  title     = {ImageNet: A Large-Scale Hierarchical Image Database},
  author    = {Deng, Jia and Dong, Wei and Socher, Richard and Li, Li-Jia
             and Li, Kai and Fei-Fei, Li},
  booktitle = {Proceedings of the IEEE Conference on Computer Vision and Pattern Recognition (CVPR)},
  pages     = {248--255},
  year      = {2009}
}

@inproceedings{loshchilov2019adamw,
  title     = {Decoupled Weight Decay Regularization},
  author    = {Loshchilov, Ilya and Hutter, Frank},
  booktitle = {International Conference on Learning Representations (ICLR)},
  year      = {2019}
}

@inproceedings{akiba2019optuna,
  title     = {Optuna: A Next-generation Hyperparameter Optimization Framework},
  author    = {Akiba, Takuya and Sano, Shotaro and Yanase, Toshihiko
             and Ohta, Takeru and Koyama, Masanori},
  booktitle = {Proceedings of the 25th ACM SIGKDD International Conference on Knowledge Discovery and Data Mining},
  pages     = {2623--2631},
  year      = {2019}
}

@article{wu2021deepbarkid,
  title   = {Deep {BarkID}: a portable tree bark identification system by knowledge distillation},
  author  = {Wu, Fanyou and Gazo, Rado and Benes, Bedrich and Haviarova, E.},
  journal = {European Journal of Forest Research},
  volume  = {140},
  number  = {6},
  year    = {2021},
  doi     = {10.1007/s10342-021-01407-7}
}

@article{warner2025environmental,
  title   = {Environmental Sensitivity in {AI} Tree Bark Detection: Identifying Key Factors for Improving Classification Accuracy},
  author  = {Warner, Charles and Wu, Fanyou and Gazo, Rado and Benes, Bedrich and Fei, Songlin},
  journal = {Algorithms},
  volume  = {18},
  number  = {7},
  pages   = {417},
  year    = {2025},
  publisher = {MDPI},
  doi     = {10.3390/a18070417}
}

@article{surendran2025forestry,
  title   = {A forestry investigation: Exploring factors behind improved tree species classification using bark images},
  author  = {Surendran, G. K. and Deekshitha and Lukac, Martin and Vybostok, J. and Mokros, Martin},
  journal = {Ecological Informatics},
  volume  = {85},
  pages   = {102932},
  year    = {2025},
  publisher = {Elsevier},
  doi     = {10.1016/j.ecoinf.2024.102932}
}

@inproceedings{robert2020barkreid,
  title     = {Tree Bark Re-identification Using a Deep-Learning Feature Descriptor},
  author    = {Robert, Martin and Dallaire, Patrick and Gigu{\`e}re, Philippe},
  booktitle = {Proceedings of the 17th Conference on Computer and Robot Vision (CRV)},
  year      = {2020},
  eprint    = {1912.03221},
  archivePrefix = {arXiv},
  note      = {arXiv:1912.03221}
}

@misc{ali2026barknetlite,
  title  = {{BarkNet-Lite}: A Lightweight Texture and Colour Network with the {BarkBD} Benchmark for Bark-Based Tree Species Recognition in {Bangladesh}},
  author = {Ali, Aroshi and Syfullah, Md. Khalid and Ahmed, Saad},
  year   = {2026},
  eprint = {2609.07600},
  archivePrefix = {arXiv},
  primaryClass  = {cs.CV},
  note   = {arXiv:2609.07600}
}

@article{remes2019bark,
  title   = {Bark recognition using novel rotationally invariant multispectral textural features},
  author  = {Reme{\v{s}}, V. and Haindl, M.},
  journal = {Pattern Recognition Letters},
  volume  = {125},
  pages   = {612--617},
  year    = {2019},
  publisher = {Elsevier},
  doi     = {10.1016/j.patrec.2019.06.027}
}

@article{fekriershad2020bark,
  title   = {Bark texture classification using improved local ternary patterns and multilayer neural network},
  author  = {Fekri-Ershad, S.},
  journal = {Expert Systems with Applications},
  volume  = {158},
  pages   = {113509},
  year    = {2020},
  publisher = {Elsevier},
  doi     = {10.1016/j.eswa.2020.113509}
}

@misc{faizal2022barkvn,
  title  = {Automated Identification of Tree Species by Bark Texture Classification Using Convolutional Neural Networks},
  author = {Faizal, Sahil},
  year   = {2022},
  eprint = {2210.09290},
  archivePrefix = {arXiv},
  primaryClass  = {cs.CV},
  note   = {arXiv:2210.09290}
}

@misc{truonghoang2020barkvn50,
  title     = {{BarkVN-50}},
  author    = {Truong Hoang, Vinh},
  year      = {2020},
  publisher = {Mendeley Data},
  doi       = {10.17632/gbt4tdmttn.1}
}

@article{wu2021woodlongitudinal,
  title   = {Wood identification based on longitudinal section images by using deep learning},
  author  = {Wu, Fanyou and Gazo, Rado and Haviarova, E. and Benes, Bedrich},
  journal = {Wood Science and Technology},
  volume  = {55},
  pages   = {553--563},
  year    = {2021},
  publisher = {Springer},
  doi     = {10.1007/s00226-021-01261-1}
}

@article{liu2024cites,
  title   = {How to discriminate wood of {CITES}-listed tree species from their look-alikes: using an attention mechanism with the {ResNet} model on an enhanced macroscopic image dataset},
  author  = {Liu, S. and Zheng, C. and Wang, J. and Lu, Y. and Yao, J. and Zou, Z. and Yin, Y. and He, T.},
  journal = {Frontiers in Plant Science},
  volume  = {15},
  pages   = {1368885},
  year    = {2024},
  publisher = {Frontiers},
  doi     = {10.3389/fpls.2024.1368885}
}

@article{ravindran2022diffuseporous,
  title   = {Towards Sustainable {North American} Wood Product Value Chains, Part {I}: Computer Vision Identification of Diffuse Porous Hardwoods},
  author  = {Ravindran, Prabu and Owens, F. C. and Wade, A. C. and Shmulsky, R. and Wiedenhoeft, Alex C.},
  journal = {Frontiers in Plant Science},
  volume  = {12},
  pages   = {758455},
  year    = {2022},
  publisher = {Frontiers},
  doi     = {10.3389/fpls.2021.758455}
}

@misc{casanova2026woodbrcross,
  title  = {{WoodBR-Cross}: A public wood-image dataset and individual-tree-disjoint benchmark for species identification},
  author = {Casanova, Daniel and Gontijo, Alexandre Bahia and Candido Junior, Arnaldo and Nisgoski, Silvana and de Paula Filho, Pedro Luiz},
  year   = {2026},
  eprint = {2610.12676},
  archivePrefix = {arXiv},
  primaryClass  = {cs.CV},
  note   = {arXiv:2610.12676}
}

@inproceedings{hafemann2014forest,
  title     = {Forest Species Recognition Using Deep Convolutional Neural Networks},
  author    = {Hafemann, Luiz G. and Oliveira, Luiz S. and Cavalin, Paulo},
  booktitle = {Proceedings of the 22nd International Conference on Pattern Recognition (ICPR)},
  pages     = {1103--1107},
  year      = {2014},
  publisher = {IEEE},
  doi       = {10.1109/ICPR.2014.199}
}

@article{wang2018revisiting,
  title   = {Revisiting multiple instance neural networks},
  author  = {Wang, Xinggang and Yan, Yongluan and Tang, Peng and Bai, Xiang and Liu, Wenyu},
  journal = {Pattern Recognition},
  volume  = {74},
  year    = {2018},
  doi     = {10.1016/j.patcog.2017.08.026}
}

@inproceedings{li2021dsmil,
  title     = {Dual-Stream Multiple Instance Learning Network for Whole Slide Image Classification with Self-Supervised Contrastive Learning},
  author    = {Li, Bin and Li, Yin and Eliceiri, Kevin W.},
  booktitle = {Proceedings of the IEEE/CVF Conference on Computer Vision and Pattern Recognition (CVPR)},
  pages     = {14318--14328},
  year      = {2021},
  doi       = {10.1109/CVPR46437.2021.01409}
}

@inproceedings{zhang2022dtfd,
  title     = {{DTFD-MIL}: Double-Tier Feature Distillation Multiple Instance Learning for Histopathology Whole Slide Image Classification},
  author    = {Zhang, Hongrun and Meng, Yanda and Zhao, Yitian and Qiao, Yihong and Yang, Xiaoyun and Coupland, Sarah E. and Zheng, Yalin},
  booktitle = {Proceedings of the IEEE/CVF Conference on Computer Vision and Pattern Recognition (CVPR)},
  pages     = {18802--18812},
  year      = {2022},
  doi       = {10.1109/CVPR52688.2022.01824}
}

@inproceedings{raff2023mil,
  title     = {Reproducibility in Multiple Instance Learning: A Case For Algorithmic Unit Tests},
  author    = {Raff, Edward and Holt, James},
  booktitle = {Advances in Neural Information Processing Systems (NeurIPS)},
  volume    = {36},
  year      = {2023},
  eprint    = {2310.17867},
  archivePrefix = {arXiv}
}

@inproceedings{kaufman2011leakage,
  title     = {Leakage in data mining: formulation, detection, and avoidance},
  author    = {Kaufman, Shachar and Rosset, Saharon and Perlich, Claudia},
  booktitle = {Proceedings of the 17th ACM SIGKDD International Conference on Knowledge Discovery and Data Mining},
  pages     = {556--563},
  year      = {2011},
  publisher = {ACM},
  doi       = {10.1145/2020408.2020496}
}

@article{roberts2017cv,
  title   = {Cross-validation strategies for data with temporal, spatial, hierarchical, or phylogenetic structure},
  author  = {Roberts, D. R. and Bahn, V. and Ciuti, S. and Boyce, M. S. and Elith, J. and Guillera-Arroita, G. and Hauenstein, S. and Lahoz-Monfort, J. J. and Schr{\"o}der, B. and Thuiller, W. and Warton, D. I. and Wintle, B. A. and Hartig, F. and Dormann, C. F.},
  journal = {Ecography},
  volume  = {40},
  number  = {8},
  pages   = {913--929},
  year    = {2017},
  doi     = {10.1111/ecog.02881}
}

@article{kattenborn2022spatial,
  title   = {Spatially autocorrelated training and validation samples inflate performance assessment of convolutional neural networks},
  author  = {Kattenborn, T. and Schiefer, F. and Frey, J. and Feilhauer, H. and Mahecha, M. D. and Dormann, C. F.},
  journal = {ISPRS Open Journal of Photogrammetry and Remote Sensing},
  pages   = {100018},
  year    = {2022},
  doi     = {10.1016/j.ophoto.2022.100018}
}

@inproceedings{bussola2021tiles,
  title     = {{AI} Slipping on Tiles: Data Leakage in Digital Pathology},
  author    = {Bussola, Nicole and Marcolini, Alessia and Maggio, Valerio and Jurman, Giuseppe and Furlanello, Cesare},
  booktitle = {Pattern Recognition. ICPR International Workshops and Challenges},
  series    = {Lecture Notes in Computer Science},
  year      = {2021},
  publisher = {Springer},
  doi       = {10.1007/978-3-030-68763-2_13},
  note      = {arXiv:1909.06539}
}

\end{document}